\documentclass[sigconf]{acmart}

\setcopyright{acmlicensed}
\copyrightyear{2018}
\acmYear{2018}
\acmDOI{XXXXXXX.XXXXXXX}
\acmConference[Conference acronym 'XX]{Make sure to enter the correct
  conference title from your rights confirmation email}{June 03--05,
  2018}{Woodstock, NY}
\acmISBN{978-1-4503-XXXX-X/2018/06}

\usepackage{url}
\usepackage{booktabs}
\usepackage{amsfonts}
\usepackage{nicefrac}
\usepackage{microtype}
\usepackage{graphicx}
\usepackage{amsmath}
\usepackage{siunitx}
\usepackage{enumitem}
\usepackage{multirow}
\usepackage[ruled,vlined]{algorithm2e}
\usepackage{subcaption}
\usepackage{wrapfig}
\usepackage{colortbl}
\usepackage{placeins}

\newcommand{\gainpos}[1]{\textcolor{blue!70!black}{\tiny\raisebox{0.25ex}{\thinspace(+#1)}}}
\newcommand{\gainzero}{\textcolor{black!55}{\tiny\raisebox{0.25ex}{\thinspace(+0.000)}}}
\newcommand{\gainneg}[1]{\textcolor{red!70!black}{\tiny\raisebox{0.25ex}{\thinspace(#1)}}}
\newcommand{\scoregain}[2]{\textbf{#1}\gainpos{#2}}
\newcommand{\scorezero}[1]{\textbf{#1}\gainzero}
\newcommand{\scoreneg}[2]{#1\gainneg{#2}}
\newcommand{\baseres}[1]{\textcolor{black!72}{#1}}
\newcommand{\methodbase}{\textcolor{black!70}{Base}}
\newcommand{\methodours}{\cellcolor{black!3}\textbf{SpecCal}}
\begin{document}

\title{SpecCal: Ambiguity-Aware Candidate Calibration for Infrared Spectrum-Based Molecular Structure Reconstruction}

\author{Yixuan Chen}
\authornote{These authors contributed equally to this work.}
\email{chenyx2123@mails.jlu.edu.cn}
\affiliation{%
  \institution{College of Computer Science and Technology, Jilin University}
  \city{Changchun}
  \country{China}}

\author{Bo Liu}
\authornotemark[1]
\email{liubo4936@gmail.com}
\affiliation{%
  \institution{School of Mathematics, Jilin University}
  \city{Changchun}
  \country{China}}

\author{Yusen Tan}
\email{ytan277@connect.hkust-gz.edu.cn}
\affiliation{%
  \institution{Information Hub, HKUST(GZ)}
  \city{Guangzhou}
  \country{China}}

\author{Guokun Yang}
\email{guokun.yang@ntu.edu.sg}
\affiliation{%
  \institution{School of Chemistry, Chemical Engineering and Biotechnology (CCEB), Nanyang Technological University}
  \city{Singapore}
  \country{Singapore}}

\author{Wenjie Du}
\email{duwenjie@mail.ustc.edu.cn}
\affiliation{%
  \institution{School of Software Engineering, University of Science and Technology of China}
  \city{Hefei}
  \country{China}}
  
\author{Jun Xia}
\correspondingauthor
\email{junxia@hkust-gz.edu.cn}
\affiliation{%
  \institution{Information Hub, HKUST(GZ)}
  \city{Guangzhou}
  \country{China}}
\affiliation{%
  \institution{HKUST}
  \city{Hong Kong SAR}
  \country{China}}

\renewcommand{\shortauthors}{Chen and Liu et al.}
\begin{abstract}
\label{abstract}
Inferring molecular structures from infrared (IR) spectra is a fundamental yet challenging problem. A key difficulty is that an IR spectrum provides limited structural information: different molecules may share similar functional groups and local vibrational patterns, leading to highly similar spectral responses. Thus, even when an observed spectrum has a unique underlying structure, reconstructing it from the spectrum remains ambiguous. Existing IR-to-molecule models usually generate a ranked set of candidate molecules, but this set is largely determined by the model's learned generation preference and may not fully capture the structures that best satisfy the observed spectral constraints. To address this limitation, we propose \textbf{SpecCal}, a training-free candidate calibration framework for IR-to-molecule prediction. SpecCal operates on the candidate outputs of existing base models and improves the prediction set by re-ranking current candidates while introducing additional structurally plausible alternatives guided by spectral consistency. The framework is plug-and-play and model-agnostic, requiring no parameter updates for integration with diverse base models. Experiments on multiple benchmarks show that SpecCal consistently improves top-$k$ reconstruction at both SMILES and scaffold levels across different base models. Further analyses demonstrate that calibrating candidate sets under spectral ambiguity provides a practical way to improve molecular reconstruction from IR spectra. The code is available at: \url{https://anonymous.4open.science/r/SpecCal-B18A}.
\end{abstract}



\maketitle

\section{Introduction}
\label{Sec: Introduction}
Infrared (IR) spectroscopy is a widely used analytical technique for molecular characterization, offering advantages such as being non-destructive, rapid, and low-cost compared to methods like nuclear magnetic resonance (NMR) and mass spectrometry (MS)~\cite{smith2011fundamentals,baker2014using,jaggi2006fourier}. These properties make IR an appealing modality for large-scale and automated molecular structure identification, motivating interest in data-driven molecular structure reconstruction from IR spectra, known as IR-to-molecule prediction~\cite{alberts2024leveraging,feng2025intelligent,devata2024deepspinn,wu2025transformer,kanakala2024spectra,shen2025molspectllm}.

Existing solutions to the IR-to-molecule task can be broadly categorized into two paradigms. The first paradigm is database search~\cite{stein1994optimization,kind2010advances}, which identifies candidate structures by comparing a query spectrum against a reference spectral library. While this approach typically achieves high accuracy, it is inherently limited by the coverage of the database.
The second paradigm leverages an encoder-decoder framework, framing IR-to-molecule as a conditional sequence-to-sequence translation task, where the IR spectrum is represented as a sequence of intensity values and the molecular structure is expressed as a SMILES sequence~\cite{weininger1988smiles}. In contrast, this approach does not rely on predefined libraries and can generalize to unseen molecules. Recent approaches employ Transformer-based sequence-to-sequence models~\cite{Transformer} to learn the mapping from IR spectra to molecular structures~\cite{alberts2024leveraging,wu2025transformer,shen2025molspectllm,zhang2025toward}. These methods have achieved strong reconstruction performance, highlighting the promise of Transformer-based models for this task.

However, existing approaches do not explicitly address the spectral ambiguity of the IR-to-molecule problem: distinct molecular structures may produce highly similar IR spectra, making structure reconstruction from a single observed spectrum difficult. To improve prediction performance, existing models typically rely on likelihood-based decoding strategies, such as beam search~\cite{vijayakumar2016diverse,freitag2017beam,he2017decoding,yang2018breaking}, to construct a ranked set of candidate structures during inference. Under spectral ambiguity, however, this likelihood-based ranking may not reliably reflect spectral consistency: high-ranked candidates can still be poorly aligned with the observed spectrum, while lower-ranked candidates may better satisfy spectral constraints.
To address this limitation, it is desirable to explore a broader set of candidate structures during inference while using spectral consistency as an explicit calibration signal for test-time candidate refinement of the prediction set. 

To this end, we propose \textbf{SpecCal}, a training-free candidate calibration framework for IR-to-molecule prediction. SpecCal operates on the candidate set produced by an existing base model and refines it by jointly considering model preference and spectral consistency. Rather than relying only on the original generated candidates, SpecCal improves the candidate set from two complementary aspects: it re-ranks existing candidates according to spectral consistency and explores additional structurally plausible candidates that may be missed by the initial base model.
Specifically, we implement this calibration process with Monte Carlo Tree Search (MCTS)~\cite{browne2012survey,coulom2006efficient}, which naturally balances exploration of new molecular candidates and exploitation of candidates with high spectral consistency. At each iteration, MCTS performs selection, expansion, evaluation, and backpropagation under spectral-consistency guidance. Through this process, MCTS jointly re-ranks existing candidates and explores new structurally plausible candidates, producing a calibrated prediction set guided by model preference and spectral consistency.
Importantly, SpecCal is plug-and-play and model-agnostic, requiring no parameter updates for integration with diverse IR-to-molecule models as the base model. By allocating additional computation during inference, it improves the quality of the final candidate set without retraining the underlying base model or changing its architecture.

Empirically, we evaluate SpecCal on two benchmark datasets with multiple IR-to-molecule base models. We report top-$k$ hit rates under both SMILES match and scaffold match to assess molecular reconstruction at different structural levels. The results show that SpecCal consistently improves candidate-set quality across evaluated datasets and base models. Further analyses examine how search scale and hyperparameter choices affect reconstruction accuracy and structural diversity.

\raggedbottom
In summary, our contributions are as follows:
\begin{itemize}[leftmargin=1.5em, topsep=2.5pt]
\item \textbf{Candidate calibration for IR-to-molecule reconstruction.} We propose SpecCal, a training-free and plug-and-play framework that calibrates the candidate set produced by an existing IR-to-molecule base model. Instead of relying only on the original generated candidates, SpecCal promotes spectrum-consistent structures within the initial set and explores additional plausible alternatives during inference.

\item \textbf{Spectral-consistency-guided search.} We implement SpecCal with Monte Carlo Tree Search (MCTS), adapting selection, expansion, evaluation, and backpropagation to molecular candidate calibration. The resulting search process uses spectral consistency as an explicit calibration signal to guide both re-ranking and structural exploration.

\item \textbf{Improved reconstruction across datasets and base models.} We demonstrate that SpecCal consistently improves top-$k$ reconstruction at both SMILES and scaffold levels on multiple benchmark datasets and base models. Ablation studies and case analyses further examine how candidate coverage, search scale, and structural diversity affect the final prediction set.
\end{itemize}

\begin{figure*}[!t]
    \centering
    \includegraphics[width=0.84\textwidth]{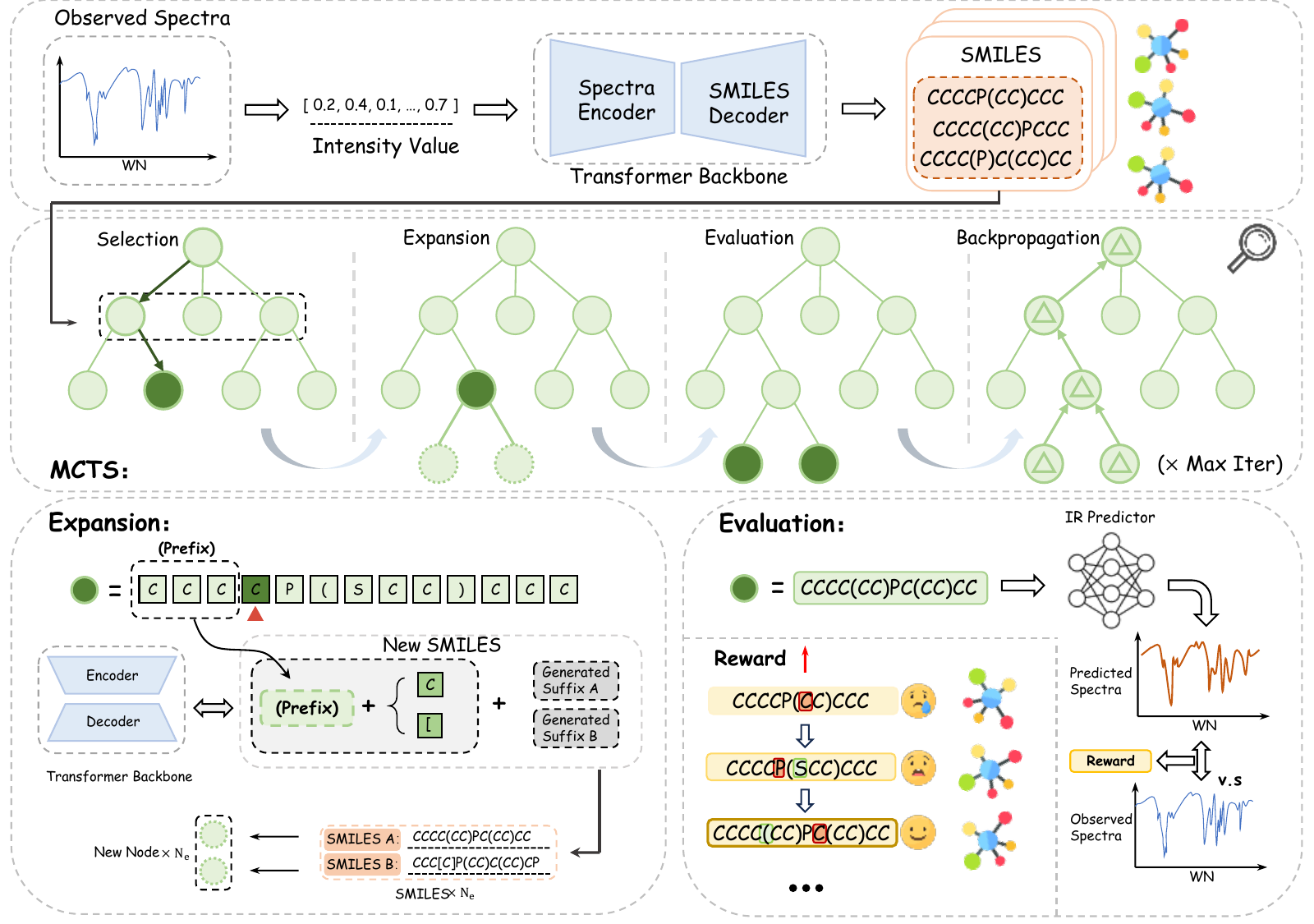} 
    \caption{Overview of SpecCal. An input IR spectrum is first translated into a set of candidate SMILES sequences by a pretrained Transformer-based base model, and these initial predictions are then calibrated by SpecCal, a training-free and plug-and-play candidate calibration framework.}
    \vspace{-5pt}  
    \label{fig:pipeline}
\end{figure*}

\section{Related Works}
\subsection{IR to Molecule}

Existing approaches to IR-to-molecule prediction can be broadly categorized into the following two paradigms: database search and encoder-decoder framework.

Database search methods match a query spectrum against reference spectral libraries to retrieve candidate molecules, and have been widely used in analytical chemistry with resources such as the NIST Chemistry WebBook~\cite{linstrom2001nist}. Recent efforts extend this paradigm by constructing large-scale in silico IR spectral libraries and performing similarity-based retrieval~\cite{kanakala2024spectra,houthuijs2023silico}. However, these approaches are fundamentally limited by database coverage and cannot identify molecules absent from the library~\cite{alberts2024leveraging,kind2010advances}, restricting their applicability to known chemical space.

Encoder-decoder frameworks formulate IR-to-molecule prediction as a sequence-to-sequence task, where the spectrum is represented as a sequence of intensities and the molecular structure is generated as a SMILES string~\cite{weininger1988smiles}. IR2Mol~\cite{alberts2024leveraging} first demonstrated the effectiveness of Transformer-based models trained on simulated spectra. Subsequent work improves spectral representation and modeling capacity, including patch-based Transformers~\cite{wu2025transformer} and molecular foundation models that jointly model spectroscopy and molecular structure~\cite{shen2025molspectllm}. Zhang and Ha~\cite{zhang2025toward} further propose a CoCa-based framework for joint spectrum--molecule alignment and sequence generation.

Despite these advances, encoder-decoder methods primarily focus on improving model performance during training, with limited attention to test-time strategies. In contrast, SpecCal is a training-free calibration framework that operates at test time, refining and expanding candidate structures to enable more comprehensive candidate discovery while promoting higher spectral consistency with the observed spectrum.

\subsection{IR Spectrum Prediction}

Complementary to IR-to-molecule generation, the forward task aims to predict IR spectra from molecular structures, which serves as a key component for evaluating spectral consistency.
Traditional approaches rely on quantum chemistry methods such as density functional theory (DFT)~\cite{becke1993density,henschel2020theoretical,katari2017improved}, which are accurate but computationally expensive and difficult to scale.
Recent advances in machine learning enable efficient IR spectrum prediction directly from molecular representations~\cite{mcgill2021predicting,gastegger2017machine,saquer2024infrared}. In particular, Chemprop-IR models molecules using a directed message passing neural network (D-MPNN)~\cite{mcgill2021predicting}, learning latent representations tailored for spectral prediction. Benefiting from pretraining on quantum chemistry data and model ensembling, it achieves strong generalization and high-quality spectral predictions.
In this work, we adopt a pretrained Chemprop-IR model as a forward spectral predictor to measure the consistency between candidate molecular structures and the observed IR spectrum.

\section{Preliminaries} 

\textbf{Problem Definition.}
The IR-to-molecule problem aims to infer the molecular structure corresponding to a given IR spectrum. 
Let $s \in \mathbb{R}^{N}$ denote the input IR spectrum represented as a sequence of absorbance values over $N$ wavenumbers, and let $y = [y_1, y_2, \dots, y_L]$ denote the target molecular representation in the form of a SMILES sequence of length $L$, where each token $y_i$ belongs to a predefined vocabulary $\mathcal{V}$.
The task can be formulated as learning a conditional mapping from spectra to molecular sequences:
\begin{equation}
f_{\theta}: s \rightarrow y,
\end{equation}
where $\theta$ denotes the model parameters. In practice, this mapping is highly ambiguous due to the limited structural information encoded in IR spectra. As a result, learning an accurate mapping $f_{\theta}$ is fundamentally challenging, and the predicted structure may not correspond to the ground truth.

\textbf{Transformer-based Generation.}
Recent approaches model IR-to-molecule prediction as a sequence-to-sequence translation task using Transformer-based architectures. 
Given an input spectrum $s$, the model generates a SMILES sequence autoregressively:
\begin{equation}
p_{\theta}(y \mid s) = \prod_{t=1}^{L} p_{\theta}(y_t \mid y_{<t}, s),
\end{equation}
where $y_{<t}$ denotes the previously generated tokens.
At inference time, decoding strategies such as beam search are typically employed to approximate the most likely sequences. However, these likelihood-driven methods tend to focus on a narrow set of high-probability candidates, which may fail to cover the true underlying structure under the inherent ambiguity of the task.



\begin{figure*}[!tbp]
    \centering
    \includegraphics[width=0.80\textwidth]{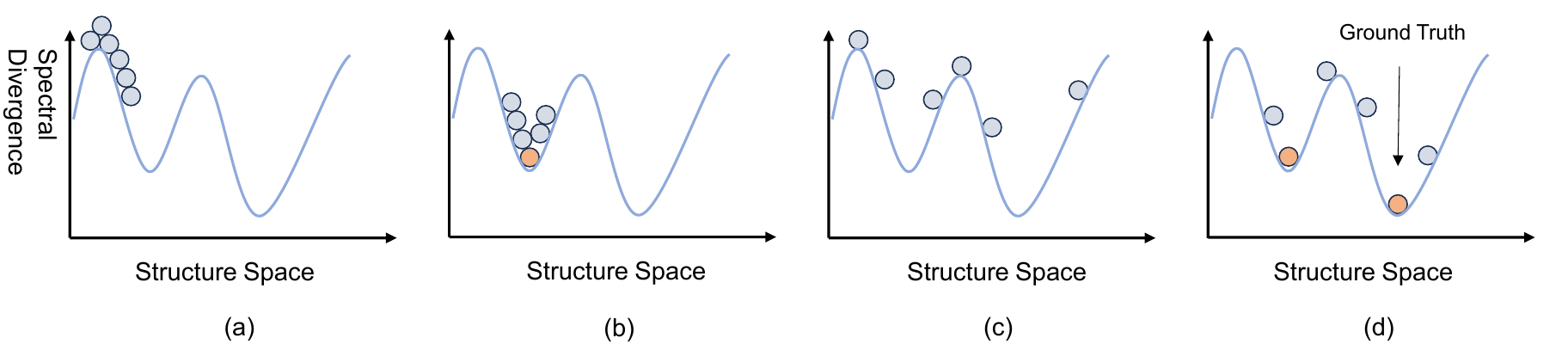}
    \caption{(a) low diversity and low spectral consistency, (b) low diversity and high spectral consistency, (c) high diversity and low spectral consistency, and (d) high diversity and high spectral consistency. Our target is the setting in (d), where the final structural candidate set maintains high consistency with the observed spectrum while preserving high structural diversity.}
    \label{fig:consitensy and diversity}
\end{figure*}
\section{Method}
\label{method}

In this section, we present SpecCal, a training-free and plug-and-play candidate calibration framework for IR-to-molecule prediction. Figure~\ref{fig:pipeline} provides an overview of the proposed calibration pipeline. Given the candidate set produced by a base model, SpecCal aims to improve the final prediction set by re-ranking existing candidates and exploring additional structurally plausible alternatives under spectral-consistency guidance. We implement this calibration process with Monte Carlo Tree Search (MCTS), where selection, expansion, evaluation, and backpropagation are adapted to search over SMILES sequences and refine molecular candidates during inference. Algorithm~\ref{alg:speccal} summarizes the overall procedure, and Figure~\ref{fig:consitensy and diversity} illustrates the calibration objective of obtaining a candidate set that better balances model preference, spectral consistency, and structural diversity under ambiguous IR spectra.

\subsection{Search Space Formulation}
\label{Search Space Formulation}

We formulate molecular refinement as a search over the space of SMILES sequences. Let $y = (y_1, \dots, y_L)$ denote a sequence of length $L$. Instead of generating $y$ from scratch, we define a local editing process that iteratively refines an initial sequence $y^{(0)}$:
\begin{equation}
y^{(t+1)} = \mathcal{E}\big(y^{(t)}, \delta^{(t)}\big),
\end{equation}
where $\delta^{(t)}$ denotes a local edit and $\mathcal{E}(\cdot)$ applies the edit followed by autoregressive completion.
Our objective is to identify a sequence that maximizes spectral consistency with the observed spectrum:
\begin{equation}
y^* = \arg\max_y \; \mathcal{S}(y, s),
\end{equation}
where $\mathcal{S}(\cdot, \cdot)$ denotes a spectral consistency score between a candidate structure and the observed IR spectrum, which serves as a guidance signal for the search process.

\subsection{Tree Search Procedure}
\label{Tree Search Procedure}

SpecCal employs Monte Carlo Tree Search to explore the space of candidate SMILES sequences in a structured and iterative manner. The initial candidate molecules generated by the Transformer serve as the root-level nodes of the search tree. Each node corresponds to a complete sequence, and edges represent stochastic refinement steps. The search iteratively performs four stages: selection, expansion, evaluation, and backpropagation.

\textbf{Selection.}
The selection process starts from the root node and recursively selects child nodes until reaching a leaf node $z^{(t)}$. At each step, the child node $z$ with the highest upper confidence bound (UCB)~\cite{kocsis2006bandit} score is selected, which balances exploration and exploitation during tree traversal:
\begin{equation}
\mathrm{UCB}(z) = Q(z) + C \sqrt{\frac{\ln N(p(z))}{N(z)}},
\label{eq:ucb}
\end{equation}
where $Q(z)$ denotes the value of node $z$, $N(z)$ is its visit count, $p(z)$ denotes its parent, and $C$ is a hyperparameter that controls the trade-off between exploitation and exploration.

\textbf{Expansion.}
Given a selected node $z^{(t)}$ with associated SMILES sequence $y(z^{(t)})$, we sample $N_e$ candidate sequences:
\begin{equation}
y_k^{(t+1)} \sim p_\theta(\cdot \mid y(z^{(t)}), s), \quad k=1,\dots,N_e,
\end{equation}
where $p_\theta$ denotes the conditional distribution predicted by the base Transformer model with parameters $\theta$. Each sampled sequence $y_k^{(t+1)}$ forms a new child node $z_k^{(t+1)}$ of $z^{(t)}$, where $y(z_k^{(t+1)}) = y_k^{(t+1)}$. The detailed expansion strategy is described in Section~\ref{Expansion Strategy}.

\textbf{Evaluation.}
Each newly expanded node $z_k^{(t+1)}$ corresponds to a SMILES sequence $y_k^{(t+1)}$, 
for which we assign a reward:
\begin{equation}
r(y_k^{(t+1)}) = \mathcal{S}(y_k^{(t+1)}, s),
\end{equation}
where $\mathcal{S}(\cdot,\cdot)$ measures the consistency between the candidate structure and the input spectrum. Details are provided in Section~\ref{Evaluation Strategy}.

\textbf{Backpropagation.}
The reward obtained at the expanded node $z_k^{(t+1)}$ is propagated along the selected path. For each node $u$ on the path from the root to $z_k^{(t+1)}$, the statistics are updated as:
\begin{equation}
N(u) \leftarrow N(u) + 1, \quad
Q(u) \leftarrow Q(u) + \frac{r(y_k^{(t+1)}) - Q(u)}{N(u)},
\end{equation}
where $r(y_k^{(t+1)})$ is the reward of the expanded sequence.

\textbf{Termination and Candidate Selection.}
The search terminates when a maximum number of iterations is reached or a high-reward node is found. 
Instead of simply selecting the top-$K$ highest-reward SMILES sequences corresponding to the explored nodes, we adopt a diversity-aware selection strategy based on Maximal Marginal Relevance (MMR)~\cite{carbonell1998use}.
Specifically, given a pool of candidate sequences, we iteratively select sequences that balance reward and diversity:
\begin{equation}
\mathrm{MMR}(y) = \lambda \cdot \hat{r}(y) + (1 - \lambda)\cdot \big(1 - \max_{y' \in \mathcal{Y}} \mathrm{sim}(y, y')\big),
\end{equation}
where $\hat{r}(y)$ is the normalized reward, $\mathcal{Y}$ is the set of already selected candidates, and $\mathrm{sim}(\cdot,\cdot)$ is the Tanimoto similarity between molecular fingerprints.
At each step, the candidate with the highest MMR score is added to $\mathcal{Y}$. The process continues until $K$ sequences are selected.
This strategy ensures that the final predictions are both high-quality and structurally diverse, which is critical for capturing the inherent ambiguity of IR-to-molecule prediction, and improving robustness across multiple plausible molecular candidates.

\begin{algorithm}[!t]
\caption{SpecCal Candidate Calibration at Test Time}
\label{alg:speccal}
\footnotesize
\KwIn{IR spectrum $s$, pretrained model $p_\theta$, initial candidate set $\mathcal{Y}_0$, target candidate size $K$, maximum iterations $T$, expansion size $N_e$, exploration constant $C$, trade-off parameter $\lambda$}
\KwOut{Refined set of molecular structures $\mathcal{Y}$}

Initialize root nodes $\mathcal{Z}_0$ using $\mathcal{Y}_0$\;
Initialize search tree $\mathcal{T}$ with $\mathcal{Z}_0$\;

\For{$t = 1$ \KwTo $T$}{
    \textbf{Selection:} Select a node $z$ from $\mathcal{T}$ using UCB:
    \[
    z \leftarrow \arg\max_{z'} \left( Q(z') + C \sqrt{\frac{\ln N(p(z'))}{N(z')}} \right)
    \]

    \textbf{Expansion:} Generate $N_e$ molecular structure candidates from $z$:
    \[
    y_k \sim p_\theta(\cdot \mid y(z), s), \quad k=1,\dots,N_e
    \]
    Create child nodes $\{z_k\}$ with $y(z_k)=y_k$ and add them to $\mathcal{T}$\;

    \textbf{Evaluation:} Compute reward for each new node:
    \[
    r_k = \mathcal{S}(y_k, s)
    \]

    \textbf{Backpropagation:} Update $Q(\cdot)$ and $N(\cdot)$ along the path from $\{z_k\}$ to the root\;
}

\textbf{Candidate Selection:}
Initialize $\mathcal{Y} = \emptyset$\;

\While{$|\mathcal{Y}| < K$}{
    Select
    \[
    z^* = \arg\max_{z \in \mathcal{T}} \left[
    \lambda \cdot \hat{r}(y(z))
    + (1-\lambda)\left(
    1 - \max_{y' \in \mathcal{Y}} \mathrm{sim}(y(z), y')
    \right)
    \right]
    \]
    Add $y(z^*)$ to $\mathcal{Y}$\;
}

\textbf{Return:} $\mathcal{Y}$\;

\end{algorithm}

\subsection{Candidate Expansion Strategy}
\label{Expansion Strategy}

We adopt an uncertainty-aware expansion strategy that refines uncertain regions of the current SMILES sequence during iterative search. Given $y^{(t)} = (y_1, \dots, y_L)$, we estimate token-level uncertainty using Shannon entropy~\cite{shannon1948mathematical} over the model's predictive distribution at each sequence position:
\begin{equation}
H_i = - \sum_{v \in \mathcal{V}} p_\theta(v \mid y_{<i}, s)\log p_\theta(v \mid y_{<i}, s),
\end{equation}
and select the most uncertain position $i^* = \arg\max_i H_i$.
At the selected position, we sample $N_c$ candidate tokens from the predictive distribution to form expansion candidates:
\begin{equation}
\tilde{y}_{i^*,k} \sim p_\theta(\cdot \mid y_{<i^*}, s), 
\quad k = 1,\dots,N_c .
\end{equation}
For each sampled token, we construct a partial sequence by preserving the prefix $y_{<i^*}$ and replacing the token at position $i^*$ with $\tilde{y}_{i^*,k}$. Each partial sequence is then completed autoregressively:
\begin{equation}
y_k^{(t+1)} \sim p_\theta(\cdot \mid y_{<i^*}, \tilde{y}_{i^*,k}, s),
\quad k = 1,\dots,N_c .
\end{equation}

This produces $N_c$ expanded SMILES candidates from the current sequence, focusing exploration on low-confidence regions while preserving high-confidence substructures.

\subsection{Spectral Evaluation and Reward}
\label{Evaluation Strategy}
To guide the search process, we design a reward function based on spectral consistency between the observed IR spectrum and that predicted from a candidate molecular structure. 
Given a candidate molecule $y$, we first obtain its corresponding predicted IR spectrum:
\begin{equation}
\hat{s} = f_{\mathrm{IR}}(y),
\end{equation}
where $f_{\mathrm{IR}}$ denotes an IR predictor that maps a molecular structure to its corresponding IR spectrum. In our implementation, we adopt a pretrained Chemprop-IR model~\cite{mcgill2021predicting}. 
We then quantify the similarity between the predicted spectrum $\hat{s}$ and the observed spectrum $s$ using Spectral Information Similarity (SIS)~\cite{mcgill2021predicting}. The similarity is computed based on spectral information divergence (SID)~\cite{chang1999spectral}:
\begin{equation}
\mathrm{SID}(\hat{s}, s) 
= \sum_i \hat{s}_i \log \frac{\hat{s}_i}{s_i} 
+ s_i \log \frac{s_i}{\hat{s}_i}.
\end{equation}

Finally, we convert this divergence into a bounded reward~\cite{mcgill2021predicting}:
\begin{equation}
\mathcal{S}(y, s) = \frac{1}{1 + \mathrm{SID}(\hat{s}, s)},
\end{equation}
which lies in $(0,1]$, with higher values indicating better spectral agreement.
This reward formulation provides a smooth and informative signal for guiding search, encouraging candidate molecules whose predicted spectra are highly consistent with the observed IR measurements. We analyze the associated computational overhead, including runtime and inference-time trade-offs, in Section~\ref{Exp analysis}.

\section{Experiments}
\begin{table*}[!t]
\centering
\caption{Top-$k$ molecular reconstruction accuracy (\%). For each base model and dataset, SpecCal denotes calibrated predictions, and blue/red values show the absolute percentage-point gain/loss over the corresponding base prediction.}
\label{tab:hit_rate_results}
\scriptsize
\setlength{\tabcolsep}{6.2pt}
\renewcommand{\arraystretch}{1.16}
\resizebox{0.96\textwidth}{!}{%
\begin{tabular}{@{}ll l ccc ccc@{}}
\toprule
\multirow{2}{*}{\textbf{Base Model}} & \multirow{2}{*}{\textbf{Dataset}} & \multirow{2}{*}{\textbf{Method}}
& \multicolumn{3}{c}{\textbf{SMILES Level}} 
& \multicolumn{3}{c}{\textbf{Scaffold Level}} \\
\cmidrule(lr){4-6} \cmidrule(l){7-9}
& & & Top-1 & Top-5 & Top-10 & Top-1 & Top-5 & Top-10 \\
\midrule

\multirow{4}{*}{\textbf{IR2Mol}}
& \multirow{2}{*}{NIST} & \methodbase & \baseres{37.115} & \baseres{58.654} & \baseres{62.500} & \baseres{81.538} & \baseres{88.077} & \baseres{89.231} \\
& & \methodours & \scoregain{54.423}{17.308} & \scoregain{62.885}{4.231} & \scoregain{63.846}{1.346} & \scoregain{85.192}{3.654} & \scoregain{90.000}{1.923} & \scoregain{90.385}{1.154} \\
\cmidrule(lr){2-9}
& \multirow{2}{*}{SynSet} & \methodbase & \baseres{63.507} & \baseres{88.152} & \baseres{91.943} & \baseres{87.204} & \baseres{95.261} & \baseres{96.209} \\
& & \methodours & \scoregain{74.882}{11.375} & \scoregain{88.626}{0.474} & \scoregain{92.417}{0.474} & \scoregain{92.891}{5.687} & \scoregain{95.735}{0.474} & \scorezero{96.209} \\
\midrule

\multirow{4}{*}{\textbf{Patch Transformer}}
& \multirow{2}{*}{NIST} & \methodbase & \baseres{47.399} & \baseres{73.410} & \textbf{78.805} & \baseres{86.898} & \baseres{92.871} & \baseres{94.220} \\
& & \methodours & \scoregain{63.654}{16.255} & \scoregain{76.154}{2.744} & \scoreneg{78.462}{-0.343} & \scoregain{88.077}{1.179} & \scoregain{93.654}{0.783} & \scoregain{94.231}{0.011} \\
\cmidrule(lr){2-9}
& \multirow{2}{*}{SynSet} & \methodbase & \baseres{68.720} & \textbf{90.995} & \baseres{94.787} & \baseres{90.995} & \baseres{95.735} & \baseres{97.630} \\
& & \methodours & \scoregain{77.725}{9.005} & \scoreneg{90.521}{-0.474} & \scorezero{94.787} & \scoregain{93.839}{2.844} & \scoregain{97.156}{1.421} & \scorezero{97.630} \\
\midrule

\multirow{4}{*}{\textbf{CoCa}}
& \multirow{2}{*}{NIST} & \methodbase & \baseres{11.538} & \baseres{17.692} & \baseres{22.692} & \baseres{68.462} & \baseres{77.115} & \baseres{78.846} \\
& & \methodours & \scoregain{26.731}{15.193} & \scoregain{30.769}{13.077} & \scoregain{30.769}{8.077} & \scoregain{77.500}{9.038} & \scoregain{84.231}{7.116} & \scoregain{84.423}{5.577} \\
\cmidrule(lr){2-9}
& \multirow{2}{*}{SynSet} & \methodbase & \baseres{21.327} & \baseres{29.384} & \baseres{36.019} & \baseres{72.512} & \baseres{81.517} & \baseres{83.886} \\
& & \methodours & \scoregain{47.867}{26.540} & \scoregain{49.289}{19.905} & \scoregain{49.289}{13.270} & \scoregain{83.412}{10.900} & \scoregain{90.047}{8.530} & \scoregain{90.521}{6.635} \\
\bottomrule
\end{tabular}%
}
\vspace{1pt}
\end{table*}

\subsection{Experimental Setup}
\label{Exp Setup}
\textbf{Dataset.}
We conduct experiments on two datasets: (1) a real-world dataset of experimentally measured gas-phase IR spectra from the NIST Chemistry WebBook~\cite{linstrom2001nist}, and (2) a large-scale synthetic dataset generated via molecular dynamics simulations~\cite{alberts2024leveraging}. 
In both datasets, each IR spectrum is represented as a discretized vector, paired with its ground-truth SMILES and molecular formula. The molecules contain atoms from \{H, C, N, O, S, P, and halogens\}, excluding charged species and stereoisomers, with 6--13 heavy atoms.
The synthetic dataset contains 456,899/50,767/126,917 samples for training/validation/test, while the NIST dataset contains 4,150/518/520 samples, respectively.

\textbf{Evaluation Metrics.}
We evaluate performance using both accuracy and diversity metrics. 
For accuracy, we report top-$k$ hit rates ($k \in \{1,5,10\}$) at the SMILES and scaffold levels~\cite{bemis1996properties}, where a hit indicates that at least one of the top-$k$ predictions matches the ground-truth SMILES or scaffold.
For diversity, we measure Tanimoto similarity~\cite{rogers2010extended} and scaffold diversity, capturing pairwise structural similarity and the diversity of chemical backbones within each predicted set.
Details of all metrics are provided in the Appendix~\ref{apen:eval_metric} for completeness and reproducibility. These metrics jointly assess quality and coverage.

\textbf{Baselines.}
We apply SpecCal as a calibration strategy on top of existing transformer-based models for IR-to-molecule prediction. 
We adopt IR2Mol~\cite{alberts2024leveraging}, Patch Transformer~\cite{wu2025transformer}, and a CoCa-based model~\cite{zhang2025toward} as base models.\pagebreak[3] IR2Mol formulates the task as a sequence-to-sequence translation from IR spectra to SMILES. Patch Transformer employs a patch-based self-attention embedding with peak-aware data augmentation, enabling robust extraction of structural information from complex spectra. The CoCa-based model integrates a contrastive captioning objective with autoregressive SMILES generation, aligning spectral and molecular representations in a shared latent space. These baselines cover different modeling choices, including standard sequence-to-sequence generation, patch-based spectral encoding, and contrastive spectrum--molecule alignment. This diversity allows us to evaluate whether SpecCal provides consistent gains as a model-agnostic calibration strategy rather than being tailored to a specific architecture.

\textbf{Implementation Details.}
\label{implementation details}
All experiments are conducted on a single NVIDIA-A40 GPU (48GB). Spectra are resampled to 400--4000 cm$^{-1}$ and normalized, with model-specific resolutions.
IR2Mol and Patch Transformer are pretrained on synthetic data and fine-tuned on NIST, while the CoCa-based model is trained on synthetic data and then trained from scratch on NIST. SpecCal is applied during inference for all models.
For SpecCal, we use $C=1.5$ throughout. IR2Mol uses 50 iterations with $N_e = 1$, Patch Transformer uses 30 iterations with $N_e = 2$, and CoCa uses 50 iterations with $N_e = 2$.
For synthetic data evaluation, we select 211 samples overlapping with NIST to obtain experimentally measured spectra.

\subsection{Main Results}
\begin{table*}[!t]
\centering
\caption{Structural diversity analysis of the top-10 predicted molecular set. Lower Tanimoto similarity and higher scaffold diversity indicate more diverse predictions.}
\label{tab:diversity_results}
\scriptsize
\setlength{\tabcolsep}{8.2pt}
\renewcommand{\arraystretch}{1.32}
\begin{tabular}{@{}llcccc@{}}
\toprule
\multirow{2}{*}{\textbf{Model}} & \multirow{2}{*}{\textbf{Method}}
& \multicolumn{2}{c}{\textbf{NIST}}
& \multicolumn{2}{c}{\textbf{SynSet}} \\
\cmidrule(lr){3-4} \cmidrule(l){5-6}
& & \shortstack{\textbf{Tanimoto}\\\textbf{Similarity}$\downarrow$}
& \shortstack{\textbf{Scaffold}\\\textbf{Diversity}$\uparrow$}
& \shortstack{\textbf{Tanimoto}\\\textbf{Similarity}$\downarrow$}
& \shortstack{\textbf{Scaffold}\\\textbf{Diversity}$\uparrow$} \\
\midrule

\multirow{2}{*}{IR2Mol}
& Base & 0.3876 & 0.1929 & 0.3967 & 0.2460 \\
& SpecCal & \textbf{0.3192} & \textbf{0.2871} & \textbf{0.3297} & \textbf{0.3322} \\
\addlinespace[3pt]
\midrule

\multirow{2}{*}{PatchTrans.}
& Base & 0.5946 & 0.1806 & 0.5253 & 0.2081 \\
& SpecCal & \textbf{0.1677} & \textbf{0.4119} & \textbf{0.1953} & \textbf{0.4592} \\
\addlinespace[3pt]
\midrule

\multirow{2}{*}{CoCa}
& Base & 0.5914 & 0.1565 & 0.6665 & 0.1943 \\
& SpecCal & \textbf{0.3026} & \textbf{0.2833} & \textbf{0.3495} & \textbf{0.3573} \\
\bottomrule
\end{tabular}
\end{table*}

\textbf{Top-$k$ Accuracy.}
We evaluate SMILES and scaffold top-$k$ ($k \in \{1,5,10\}$) hit rates on both the synthetic (SynSet) and NIST datasets, as summarized in Table~\ref{tab:hit_rate_results}.
Applying SpecCal on different base models consistently improves performance, with the strongest gains concentrated at top-1 and top-5. For IR2Mol, SMILES top-1 accuracy increases from 37.1\% to 54.4\% on NIST and from 63.5\% to 74.8\% on SynSet. For Patch Transformer, SpecCal improves top-1 accuracy from 47.4\% to 63.7\% on NIST and from 68.7\% to 77.7\% on SynSet. CoCa shows the same reranking effect, with top-1 accuracy rising from 11.5\% to 26.7\% on NIST and from 21.3\% to 47.9\% on SynSet. These improvements suggest that SpecCal is not merely expanding the candidate pool, but is effectively reranking spectrum-consistent structures toward the top of the list.

For top-10, the gains are more modest for IR2Mol and Patch Transformer because the base models already retrieve a large fraction of plausible candidates within the broader beam. CoCa shows a more visible top-10 improvement, with SMILES accuracy rising from 36.0\% to 49.3\% on SynSet and from 22.7\% to 30.8\% on NIST. This suggests that SpecCal can do more than rerank existing high-confidence predictions: it can surface better structures that were not originally prioritized by the base model. The gains are also reflected at the scaffold level, showing that SpecCal improves molecular-backbone reconstruction beyond exact SMILES matching.


\begin{figure}[!t]
    \centering
    \includegraphics[width=0.96\linewidth]{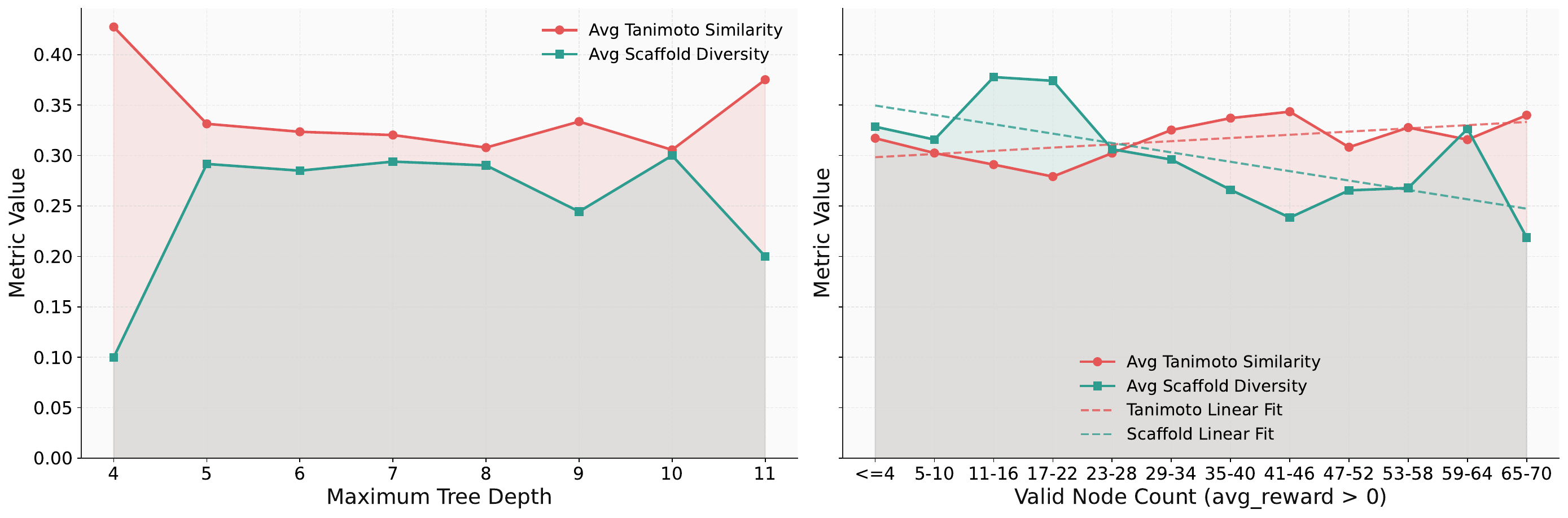}
    \caption{Bucket-wise analysis of Tanimoto similarity and scaffold diversity. The left figure shows the relationship between maximum tree depth and the diversity metrics, while the right figure illustrates the effect of the number of valid nodes (\texttt{reward > 0}). Dashed lines indicate linear trend fits, and each point represents the bucket-average over all samples within each bucket.}
    \label{fig:diversity_stat}
\end{figure}

\textbf{Diversity Evaluation of Predicted Molecules.}
We evaluate the predicted molecules of various baselines on both the synthetic SynSet and the real NIST datasets by computing the average Tanimoto similarity and scaffold diversity, as shown in Table~\ref{tab:diversity_results}. A more diverse set of predictions is expected to exhibit lower Tanimoto similarity and higher scaffold diversity. The results show that, across different base models, integrating SpecCal significantly reduces Tanimoto similarity while increasing scaffold diversity, indicating enhanced structural diversity.
Figure~\ref{fig:diversity_stat} further illustrates the relationship between search tree scale (i.e., maximum depth and the number of valid nodes) and diversity. Increasing the tree depth from 4 to 5 leads to a substantial improvement, after which the metrics stabilize. In addition, the number of valid nodes is positively correlated with diversity, highlighting the role of search scale in promoting diverse candidate generation. This observation further suggests that calibration expands the candidate set beyond near-duplicate structures.

\FloatBarrier

\subsection{SpecCal Analysis}
\label{Exp analysis}
\begin{figure}[!htbp]
    \centering
    \includegraphics[width=0.99\linewidth]{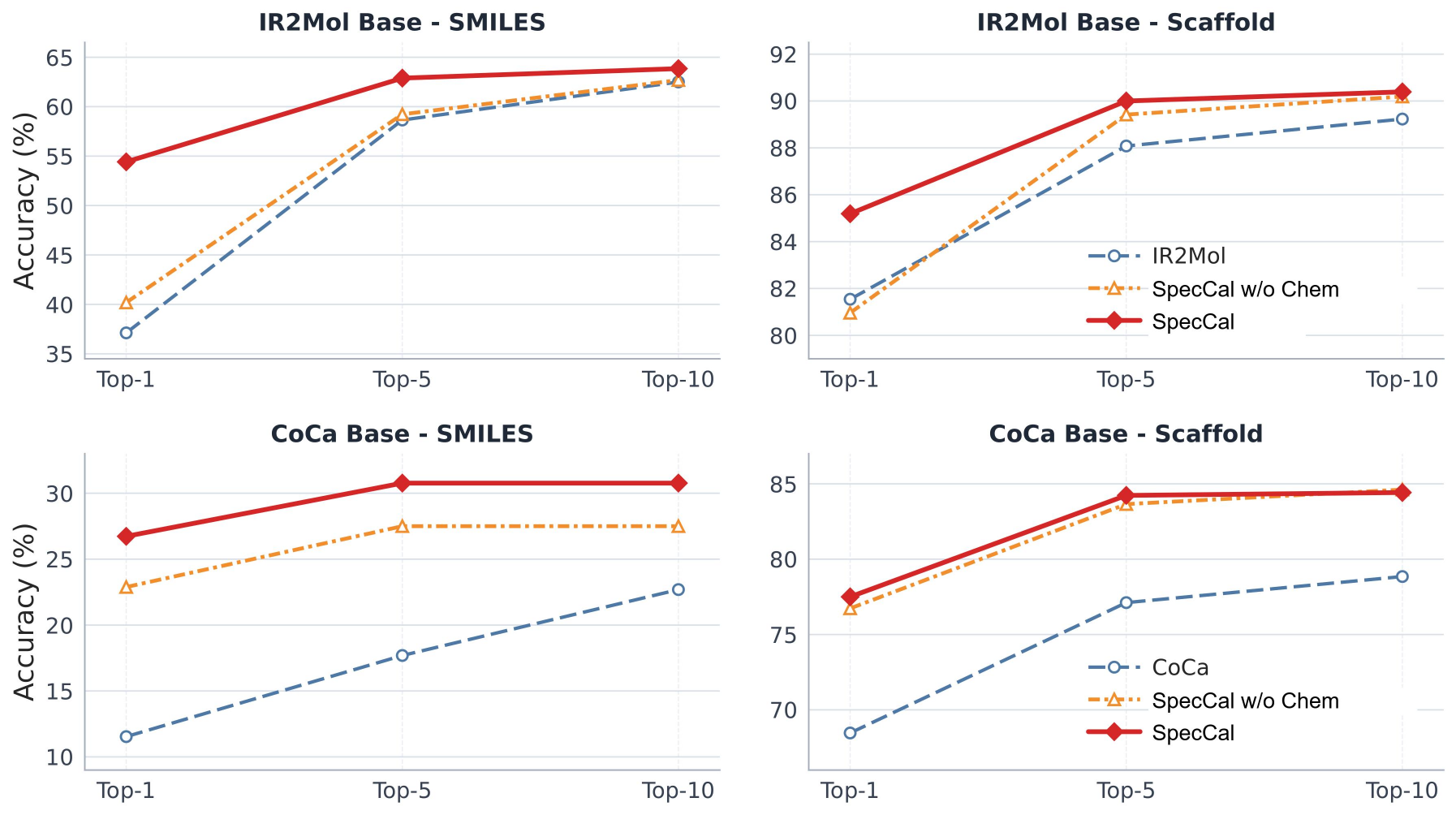}
    \caption{Ablation study on the NIST dataset. Each plot shows the top-$k$ hit rates ($k=1,5,10$) for three settings: the base model, the base model with SpecCal, and the base model with SpecCal without Chemprop-IR. The top row corresponds to IR2Mol and the bottom row to CoCa, with SMILES match (left) and scaffold match (right).}
    \label{fig:ablation}
\end{figure}

\textbf{Ablation Analysis.}
We conduct ablation experiments on the NIST dataset using IR2Mol and CoCa as base models. The evaluated component is the Chemprop-IR model used for node scoring in SpecCal. In the ablated version, Chemprop-IR is replaced with a simple chemical formula matching score; detailed reward computation is provided in the Appendix~\ref{reward_comp_details}. We report top-1, top-5, and top-10 hit rates at both SMILES and scaffold levels (Figure~\ref{fig:ablation}). The results show that the pretrained Chemprop-IR model plays a critical role in the overall performance of SpecCal, while SpecCal still yields noticeable improvements even without a dedicated IR predictor. The formula-based reward still provides a coarse chemical constraint by favoring candidates with compatible elemental composition, which explains why the ablated variant remains better than the base model. However, Chemprop-IR provides a finer spectral constraint and can better distinguish candidates with similar formulas but different structures, leading to stronger calibrated predictions.

\FloatBarrier
  
\begin{figure}[!htbp]
    \centering
    \includegraphics[width=0.99\linewidth]{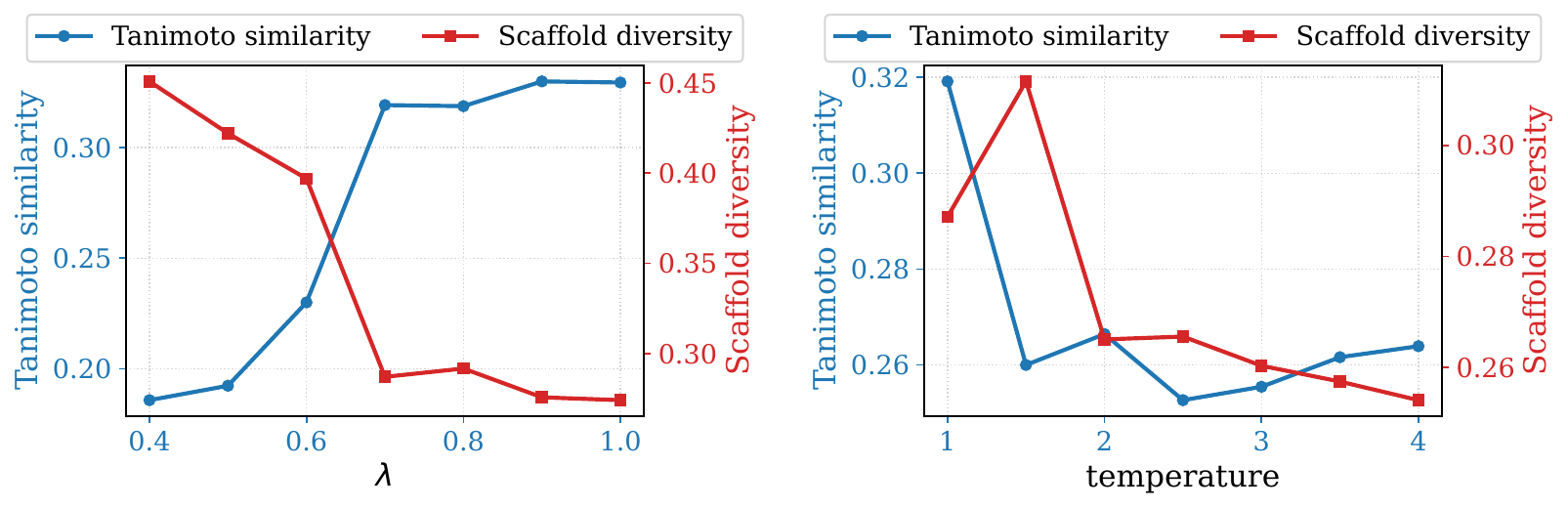}
    \caption{Sensitivity analysis of SpecCal with respect to key hyperparameters. The left panel shows the effect of $\lambda$ on diversity metrics, while the right panel illustrates the impact of temperature during the expansion stage.}
    \label{fig:sensitivity}
\end{figure}

\textbf{Sensitivity Analysis.}
\label{sensitivity ana}
We analyze the sensitivity of SpecCal on the NIST dataset using IR2Mol as the base model, focusing on $\lambda$ and temperature (Figure~\ref{fig:sensitivity}). The parameter $\lambda$ controls node selection after MCTS, while temperature governs resampling during expansion.
As $\lambda$ increases, diversity decreases, reflected by higher Tanimoto similarity and lower scaffold diversity. This is consistent with $\lambda$ controlling the balance between reward maximization and diversity promotion: smaller values favor broader exploration, whereas larger values prioritize spectrum-consistent candidates. Temperature shows a non-monotonic effect on diversity. As temperature increases, Tanimoto similarity first decreases and then increases when the temperature exceeds 2.5; scaffold diversity first improves and then declines when the temperature exceeds 1.5. These trends indicate that moderate temperature encourages exploration and produces more diverse valid candidates, whereas overly high temperature increases the probability of generating invalid structures during expansion. Since these invalid candidates are filtered out, the effective candidate set becomes less diverse. In practice, the results indicate that SpecCal is reasonably robust over a range of settings, while intermediate values of $\lambda$ and temperature offer a better balance between accuracy and diversity.

\textbf{Computational Overhead.}
SpecCal incurs additional computational cost at test time due to the iterative tree search process, including candidate expansion and spectral evaluation. Compared with standard single-pass decoding, this leads to increased inference time. However, SpecCal is training-free and requires no parameter updates, so the additional cost is incurred only during inference and avoids the substantial overhead associated with training or fine-tuning a new model. In practice, this provides a favorable trade-off, as improved reconstruction accuracy and structural diversity can be obtained without retraining. The computational cost can also be flexibly controlled by adjusting the search scale, such as the number of iterations and expansion size, allowing users to balance efficiency and performance.

\begin{figure}[!htbp]
    \centering
    \includegraphics[width=0.95\linewidth]{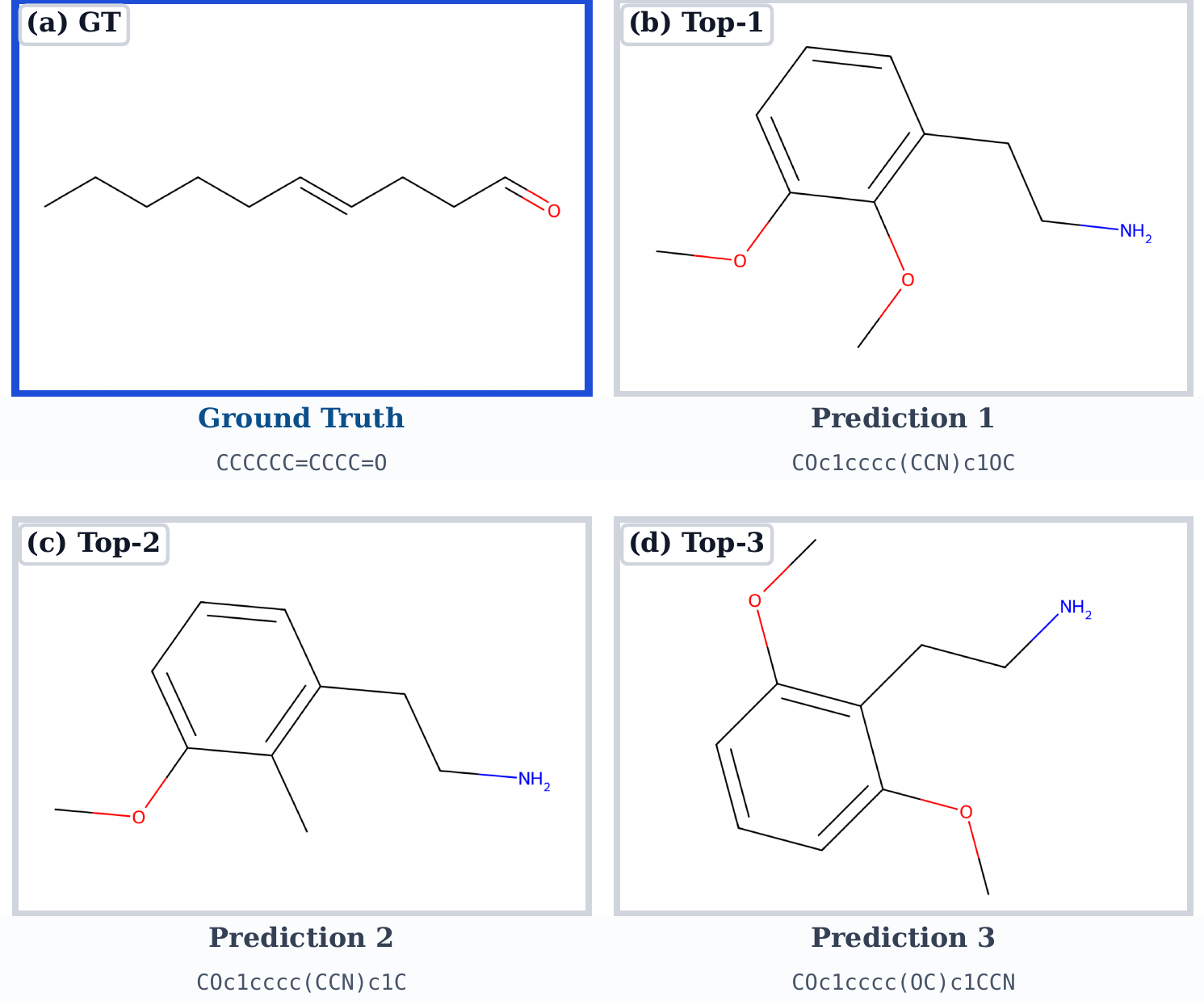}
    \caption{Top-3 predictions from the base model. The ground-truth structure is not included in the top-10 predictions, illustrating that the initial beam fails to reconstruct the correct molecule under this challenging spectrum.}
    \label{fig:case_1}
\end{figure}

\textbf{Case Study.}
We analyze a sample from the NIST dataset with ground-truth SMILES \texttt{CCCCCC=CCCC=O}. This case illustrates the second role of SpecCal discussed in Section~\ref{Sec: Introduction}: beyond re-ranking the existing candidate set, SpecCal can introduce additional structurally plausible molecules that are missed by the initial base model. As shown in Figure~\ref{fig:case_1}, the base model with beam search does not reconstruct the correct molecular structure within its top-10 predictions. The ground-truth structure is therefore absent from the original candidate set, meaning that re-ranking alone would be insufficient for correcting this prediction.

\begin{figure}[!htbp]
    \centering
    \includegraphics[width=0.62\linewidth]{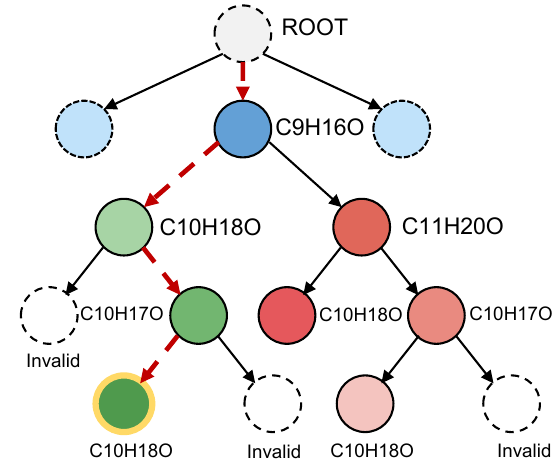}
    \caption{Partial visualization of the SpecCal calibration process for the case study. The ground-truth structure is incorporated into the calibrated candidate set and ranked among the top-10 predictions.}
    \label{fig:case_2}
\end{figure}

After applying SpecCal, the ground-truth structure is incorporated into the calibrated top-10 candidate set through test-time calibration (Figure~\ref{fig:case_2}). The corresponding node obtains a reward of \textbf{0.29} and is ranked \textbf{5th} in the refined set, indicating that spectral-consistency-guided calibration can introduce a valid structure that was not initially generated by the base model. Meanwhile, the average Tanimoto similarity decreases from \textbf{0.53} to \textbf{0.38}, suggesting that SpecCal does not simply generate minor variants of the original beam candidates, but expands the candidate set toward more diverse molecular structures. This example highlights how candidate calibration can improve IR-to-molecule reconstruction by expanding the calibrated candidate set under spectral constraints.

\FloatBarrier

\section{Conclusion}

We propose SpecCal, a training-free, plug-and-play, reward-guided calibration framework for IR-based molecular structure reconstruction. Built on Monte Carlo Tree Search, it refines Transformer-based model predictions at test time by balancing exploration of diverse candidate structures and exploitation of structures with high spectral consistency, without requiring any modification of model parameters. Experiments on multiple datasets show that SpecCal achieves state-of-the-art accuracy while improving the diversity of predicted structures, uncovering plausible molecular structures beyond the ground truth that remain highly consistent with the observed spectrum, and demonstrating the effectiveness of test-time calibration for ambiguity-aware molecular reconstruction.

\section{Limitations and Ethical Considerations}
\label{limitation}
Despite its effectiveness, SpecCal introduces additional computational overhead during inference because the search procedure is performed on a per-sample basis and requires iterative expansion. Future work will focus on improving search efficiency through adaptive budget allocation and on extending SpecCal to related spectroscopy tasks for molecular reconstruction. This work uses public or synthetic molecular spectroscopy datasets and does not involve human subjects, personally identifiable information, or private data. Potential risks mainly concern misuse for unsafe chemical design, so practical deployment should follow applicable laboratory safety, chemical compliance, and responsible use guidelines.

\section{Generative AI Usage}
Generative AI tools were used only to assist with language polishing, formatting, and clarity checking during manuscript preparation. All technical claims, experimental results, tables, figures, citations, and scientific interpretations were reviewed and verified by the authors. No generative AI system was used to produce experimental measurements, alter reported results, or make autonomous scientific decisions.

\section{Data Availability}
\label{sec:data_availability}
The datasets used in this work are publicly available. The NIST Chemistry WebBook can be accessed at \url{https://webbook.nist.gov/chemistry/}. For the synthetic dataset (SynSet), we use the dataset introduced in prior work~\cite{alberts2024leveraging}. The dataset is publicly available at \url{https://zenodo.org/records/7928396}.

\bibliographystyle{ACM-Reference-Format}
\bibliography{reference}


\appendix

\section{Evaluation Metrics}
\label{apen:eval_metric}

To evaluate the structural diversity of the predicted molecular sets, we adopt two complementary metrics: Tanimoto similarity and scaffold diversity.

\textbf{Tanimoto Similarity.}
Given a set of predicted molecules $\mathcal{Y} = \{y_1, \dots, y_n\}$, we compute the average pairwise Tanimoto similarity based on molecular fingerprints:
\begin{equation}
\mathrm{Sim}(\mathcal{Y}) = \frac{2}{n(n-1)} \sum_{i < j} \mathrm{Tanimoto}(y_i, y_j),
\end{equation}
where $\mathrm{Tanimoto}(y_i, y_j)$ denotes the Tanimoto similarity between the fingerprints of molecules $y_i$ and $y_j$. A lower value indicates higher structural diversity within the set.

\textbf{Scaffold Diversity.}
We further measure diversity at the scaffold level using Bemis--Murcko scaffolds. Given the same set $\mathcal{Y}$, scaffold diversity is defined as:
\begin{equation}
\mathrm{Div}_{\mathrm{scaffold}}(\mathcal{Y}) = \frac{|\mathcal{S}(\mathcal{Y})|}{|\mathcal{Y}|},
\end{equation}
where $\mathcal{S}(\mathcal{Y})$ denotes the set of unique scaffolds extracted from $\mathcal{Y}$. A higher value indicates a more diverse set of molecular backbones.

\textbf{Group-wise Aggregation.}
In practice, predicted molecules are organized into groups (e.g., top-$k$ candidates for each sample). The above metrics are first computed within each group and then averaged across all groups to obtain the final reported values.

\section{Reward Computation Details}
\label{reward_comp_details}
In the ablated setting, where Chemprop-IR is not used, we adopt a simple chemical formula matching score as the reward function.
Given a target molecular formula, we extract the count of each element $t_e$, and for a candidate molecule (parsed from SMILES), we obtain the corresponding element count $c_e$. Let $\mathcal{E}$ denote the set of all elements.

We first compute the $L_1$ mismatch between the candidate and target formulas:
\begin{equation}
\mathrm{mismatch}_{L_1} = \sum_{e \in \mathcal{E}} \left| c_e - t_e \right|.
\end{equation}

To normalize this value, we define:
\begin{equation}
T = \sum_{e \in \mathcal{E}} t_e, \quad
\mathrm{mismatch}_{\mathrm{norm}} = \frac{\mathrm{mismatch}_{L_1}}{\max(1, T)}.
\end{equation}

The final reward is computed as:
\begin{equation}
r_{\mathrm{nochem}} = \max\left(0, \, 1 - \mathrm{mismatch}_{\mathrm{norm}} \right).
\end{equation}

This formulation assigns a reward of $1$ when the elemental composition exactly matches the target, and gradually decreases as the mismatch increases, with a lower bound of $0$.

\section{Test-Time Scaling}
Test-Time Scaling (TTS) improves the performance of pretrained models by allocating additional computation during inference, without modifying model parameters or retraining~\cite{zhang2025survey,chen2024expanding,snell2024scaling}. It is widely used in large language models (LLMs) to explore alternative outputs or reasoning trajectories, and is particularly effective for tasks with intrinsic uncertainty or multiple plausible solutions.

TTS encompasses a range of inference-time strategies, including random sampling~\cite{wang2022self}, self-consistency, and tree-search-based methods~\cite{hao2023reasoning,yao2023tree,hooper2025ets}, all of which aim to better explore the space of candidate solutions beyond single-pass generation. Among these, tree-search methods provide a structured way to balance exploration and exploitation during inference. A representative approach is Monte Carlo Tree Search~\cite{browne2012survey,coulom2006efficient,silver2016mastering}, which incrementally builds a search tree over candidate solutions and uses a reward signal to guide the search toward promising regions.
By enabling systematic exploration of the solution space, TTS methods, especially tree-search-based approaches, can improve prediction reliability and robustness. SpecCal can be viewed as a TTS strategy for IR-to-molecule reconstruction, since it allocates additional inference-time computation to calibrate the candidate set without updating the base model.


\newpage

\end{document}